\documentclass[letterpaper, 10 pt, conference]{ieeeconf}  

\usepackage{graphicx}
\usepackage{amsmath}
\usepackage{multicol}
\usepackage{array}
\usepackage{float}
\newcolumntype{P}[1]{>{\centering\arraybackslash}p{#1}}
\usepackage{soul}

\makeatletter
\let\NAT@parse\undefined
\makeatother
\usepackage[numbers]{natbib}

\usepackage{todonotes}

\renewcommand{\baselinestretch}{0.96}

\title{\Large \bf Motion-based Object Segmentation 
based on Dense RGB-D Scene Flow}

\author{ Lin Shao, Parth Shah$^*$, Vikranth Dwaracherla$^*$, and Jeannette Bohg
\thanks{$^*$The authors contributed equally.}
\thanks{The authors are with Stanford University, CA, USA.
        {\tt\small [lins2,pshah9,vikranth,bohg]@stanford.edu}}%
}

\IEEEoverridecommandlockouts                              

\begin{document}


%



\maketitle

\begin{abstract}
Given two consecutive RGB-D images, we propose a model that estimates
a dense 3D motion field, also known as {\em scene flow\/}. We take
advantage of the fact that in robot manipulation scenarios, scenes
often consist of a set of rigidly moving objects. Our model jointly
estimates (i) the segmentation of the scene into an unknown but finite
number of objects, (ii) the motion trajectories of these objects and
(iii) the object scene flow. We employ an hourglass, deep neural
network architecture. In the encoding stage, the RGB and depth images
undergo spatial compression and correlation. In the decoding stage,
the model outputs three images containing a per-pixel estimate of the
corresponding object center as well as object translation and
rotation. This forms the basis for inferring the object segmentation
and final object scene flow. To evaluate our model, we generated a new
and challenging, large-scale, synthetic dataset that is specifically
targeted at robotic manipulation: It contains a large number of scenes
with a very diverse set of simultaneously moving 3D objects and is
recorded with a simulated, static RGB-D camera. In quantitative
experiments, we show that we outperform state-of-the-art
scene flow and motion-segmentation methods on this data set. In qualitative experiments, we show how our learned model transfers to challenging
real-world scenes, visually generating better results
than existing methods.  

\end{abstract}


\section{Introduction}
Semantic and functional scene understanding is a crucial capability of
manipulation robots. In the Computer Vision community, this
challenging problem is often approached given only a single
image. However, a robot is able to physically interact with the
environment and thereby autonomously induce motion in the scene. This
motion creates a rich, visual sensory signal that would otherwise not
be present, thus facilitating better scene understanding. Methods that
exploit physical interaction to ease perception are often referred to as
performing \textit{Interactive Perception} (IP)~\citep{Bohg:IP:17}. In this
paper, we are providing the robot with a model to process the visual
effect of its interaction. Given two consecutive RGB-D images, we are
interested in estimating a dense 3D motion field of the environment,
also known as {\em scene flow}. We show how this result helps to
segment the finite, but unknown number of moving objects in the
scene. This can provide input to tasks such as for example grasp
planning or 3D object reconstruction. 

\begin{figure}[t!]
\centering
\includegraphics[width=0.94\linewidth]{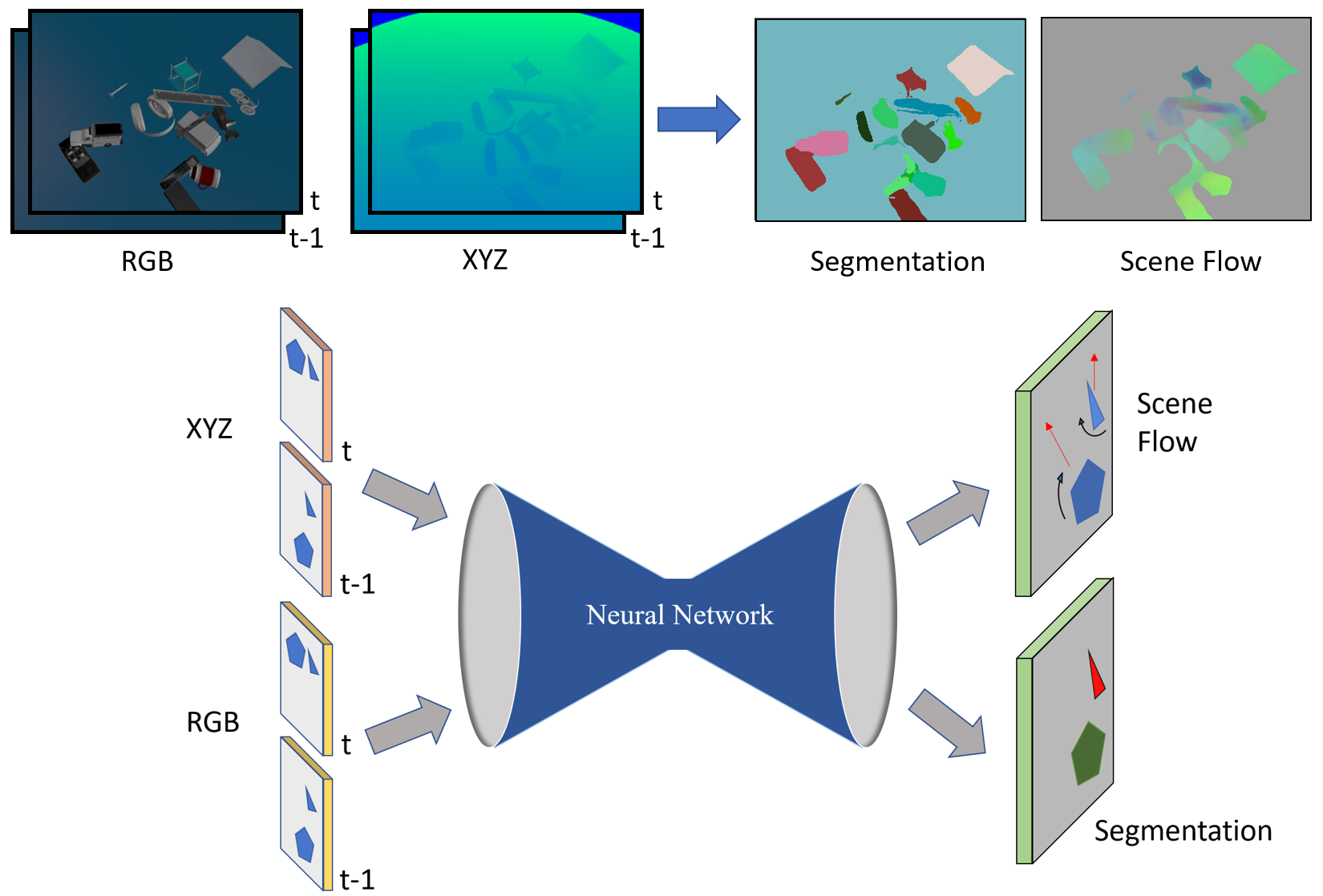}
\caption{We present a neural network which learns to estimate object
segmentation and scene flow given a pair of RGB-D images. The data
undergoes spatial compression, correlation, and refinement to propose
object segmentations and transformations.} 
\label{fig:teaser}
\end{figure}

We propose a model that takes advantage of the fact that in a common
household scenario, scenes often consist of a set of rigidly moving
objects. Our model jointly estimates (i) the segmentation of a scene
into a finite number of rigidly moving object, (ii) the motion
trajectories of these objects and (iii) the resulting {\em object
scene flow\/}~\citep{menze2015object}. We propose to use
a deep neural network architecture that takes as input a pair of
consecutive RGB-D images. See Fig.~\ref{fig:teaser} for an overview of
the approach. In a first stage, features are extracted from each of
the four input images. The RGB features are then correlated and the
resulting values are used to weight the feature encoding of the depth
data. Intuitively, this favors correspondences between points in the
depth data that also have a strong similarity in the RGB images. The
result is then decoded to produce three images containing the object
positions, their translation, and their rotation. From this, we can
infer the object scene flow and segmentation.


Our primary contributions are: (1) generating a
challenging, large-scale dataset for scene flow estimation with
ground-truth annotated RGB-D images, (2) treating rotational symmetry
of objects in scene flow prediction, (3) estimating object scene flow
with a deep neural network architecture, and (4) predicting rigid body
transformations to segment a finite, but unknown number of moving
objects. 


\section{Related Work}
Estimating scene flow has a long-standing history in the research
community starting with \citet{vedula2005three}. 
We briefly review the most recent approaches that are
related to our work in terms of several aspects: input sensor, data
sets, learning-based methods and motion segmentation.  

\subsection{Scene Flow based on RGB-D or Stereo Images}
\citet{gottfried2011computing} were the first to use an RGB-D sensor
for scene flow estimation. Their work also addresses the necessary
calibration process. \citet{herbst2013rgb} generalize the two-frame
variational optical flow algorithm (2D) to scene flow (3D). The
resulting dense scene flow is then used for rigid motion
segmentation. \citet{jaimez2015primal} present the first real-time
method for computing dense scene flow from RGB-D images. Their method
is based on a variational formulation that imposes brightness and
geometric consistencies. The minimization problem is efficiently
solved with a GPU and a primal-dual algorithm. \citet{6751281} were
the first to propose the estimation of piecewise rigid scene flow where
oversegmentation into superpixels constrains the scene flow
estimation. The authors obtain a new level of accuracy that may run in
real-time. Inspired by this
work, \citet{DBLP:journals/corr/abs-1710-02124} propose a multi-frame
scene flow approach which jointly optimizes the consistency of the
patch appearances and their local motions from RGB-D image
sequences. However the reliance on bottom up cues for segmentation may
lead to oversegmentation of objects. \citet{menze2015object} defined
{\em object scene flow\/} as the 3D motion associated with a set of
pixels that constitute a rigidly-moving object. By assuming that the
scene consists of a set of such objects and encouraging superpixels in
the same region to have similar 3D motion, the authors constrain the
solution space for estimating scene flow. The inference process is
computationally very expensive, taking 2-50 minutes per image pair.

For computing a matching score
between pixels across stereo frames and over time, traditional
approaches often rely on assumptions like brightness constancy and
motion smoothness within a small region. In real scenes, these
assumptions are often broken for example with non-Lambertian surfaces,
occlusions or large displacements. These effects are prevalent when
multiple objects are moving fast and simultaneously over
time. Therefore, matching pixel positions over time is the most
vulnerable component in traditional methods. Our hypothesis is that
these challenges can be mitigated by using 
methods that learn powerful features of the raw input
data over multiple spatial scales. Evidence comes from successful
learning-based approaches towards optical flow as detailed in 
Sec.~\ref{sec:learning_flow}. 

\subsection{Datasets}
Several large scale datasets exist for benchmarking and learning
optical and scene flow. Different from our data set, they are all
under a binocular setting with flow and disparity ground
truth. KITTI~\citep{Geiger2013IJRR} consists of 194 training and 195
test scenes recorded from a calibrated pair of cameras mounted on a
car. Ground truth annotations are obtained by combining data from a 3D
laser scanner with the car's ego motion. \citet{menze2015object}
annotated the dynamic scenes with 3D CAD models for all moving
vehicles and modified the dataset with 200 training scenes and 200
test scenes. KITTI contains valuable real world data. However, the ground
truth contains some approximation error.  
\citet{mayer2016large} created a synthetic dataset called
FlyingThings3D containing over 35000 stereo frames with ground truth
scene flow annotations. When using data from stereo cameras,
insufficient texture can result in matching errors across frames and over
time. RGB-D cameras deliver
dense depth measurements despite a lack of texture. This data can
support the matching process. Therefore, our data set contains pairs
of consecutive RGB-D images and is of similar size as FlyingThings3D. 
Different from the aforementioned datasets, objects in our dataset
are falling onto a surface, colliding with each others, and even sliding
on the surface. It is important for a manipulation robot to understand this type of non-smooth, physically-realistic motion
due to contact. The
objects in our scenes are also much closer to each other, leading to more
challenging occlusions and motion. And lastly, we use a new annotation
method to coherently label objects with rotational symmetry. See
Section~\ref{sec:dataset} for more details on our dataset.

\subsection{Learning-Based Flow Prediction}\label{sec:learning_flow}
Learning-based methods have up till now been mainly applied to optical
flow estimation. \citet{7410673} posed this problem as a supervised
learning problem and were the first to solve it with {\em
Convolutional Neural Networks\/} (CNNs). They compare two
architectures called FlownetS and FlownetC: a generic architecture and
an architecture that includes a layer that correlates feature vectors
at different image locations. These two FlowNets were tested on
datasets like Sintel~\citep{butler2012naturalistic} and
KITTI~\citep{Geiger2013IJRR} achieving competitive accuracy at frame
rates of 5-10 fps. \citet{DBLP:journals/corr/IlgMSKDB16} extend
FlowNet by developing a stacked architecture. It includes warping of
the second images with intermediate optical flow. The authors also propose a
subnetwork specializing in small displacements resulting in
state-of-the-art results while running at real-time. For
learning-based scene flow estimation, \citet{hadfield2014scene}
introduced a novel cost function. In this new formulation, only a
limited portion of the parameters from the entire pipeline are
learned, leading to limited improvements.  
\citet{mayer2016large} utilized a CNN to estimate scene flow based on
stereo images. They embed a disparity estimation network called
DispNet into FlowNet~\citep{7410673}. 
We propose an hourglass deep architecture that uses two RGB-D frames
as input. It adopts the correlation layer of FlowNetC for the
RGB encoding and uses this to associate encoded point cloud features. One of our main contributions is the decoder which directly predicts object position, translation and rotation. From this we can infer object scene flow and motion-based, rigid object segmentation.

\subsection{Motion-based Segmentation}
\citet{Bohg:IP:17} extensively review the variety of work towards
motion-based segmentation within robotics. Here, we discuss a few
representative examples. Many works use over-segmentations and connect
superpixels over time using clustering
methods~\citep{brendel2009video,grundmann2010efficient}. However, the
reliance on bottom-up cues often results in some remaining
oversegmentation.
The authors of \citep{cheriyadat2009non, brox2010object,zografos2014fast} 
formulate the problem as clustering of point trajectories across
different frames and solve it based on spectral clustering
methods. Instead, \citet{rahmati2014motion} utilize multi-label graph
cuts. \citet{ji2014robust} define an unbalanced energy to model both,
motion segmentation and point matching. \citet{KB15b} formulate
motion-based segmentation based on point trajectories as a minimum
cost, multi-cut problem. The minimum cost multi-cut formulation allows
for varying cluster sizes.  
We propose a model where each pixel directly predicts the center and
trajectory of the object that it is associated with. We achieve
accurate motion-based segmentation by clustering in this space. This
in turn helps to refine the scene flow estimate.

\section{Problem Formulation \& Notation} 
The input
to the proposed model are two consecutive RGB-D images. We assume that
the environment consists of a finite, but unknown, number of rigidly
moving objects. The network outputs (i) a pixel-wise segmentation of
each object, (ii) the rigid body motion of each object, and (iii) the
scene flow of each pixel in a reference frame. 
 
More formally, let $\mathcal{I}^{t}$ and $\mathcal{P}^{t}$ denote an RGB image and a point cloud from a single RGB-D image at time $t$. Time $t$ and $t-1$ refer to the current and previous frames, respectively.
To calculate scene flow of each point $P^t_i \in \mathcal{P}^{t}$ in a reference frame, we predict its 3D displacement by estimating its corresponding position $P^{t-1}_i$ in the previous frame. This estimate is denoted by  $\hat{P}^{t-1}_i$. 

Let $\mathcal{O}$ denote the set of rigidly moving objects in the scene. The rigid body motion between two consecutive frames for $\mathcal{O}_k$ is described by an SE(3) transform consisting of a rotation $R_k$ and translation $T_k$. 
Our model directly outputs three images $\mathcal{Q}$, $\mathcal{T}$ and $\mathcal{X}$ where each pixel contains an estimate of the rotation, translation and center of the object that the pixel belongs to. 
Therefore, if point $P^t_i$ is generated by $\mathcal{O}_k$ then the correct value at the projected image coordinates $(u,v)$ in the respective output images will contain the ground truth rotation, translation and center of object $\mathcal{O}_k$.

We denote the rotation of a point $P_i$ based on the axis-angle representation $Q_k$ as $r(P_i,Q_k) = R_kP_i$. Therefore, the corresponding point in frame $t-1$ can be computed by 
\begin{equation}\label{eq:sceneflow}
P^{t-1}_i=r(P^{t}_i-X_k,Q_k)+X_k+T_{k}
\end{equation} with per-pixel scene flow $S_i=P^{t-1}_i-P^{t}_i$. 
Note, that our model outputs an estimate of the ground truth variables $Q_k, T_k$ and $X_k$ which results in $\hat{P}^{t-1}_i$ instead of $P^{t-1}_i$ and therefore only in an estimate $\hat{S}_i$ of the ground truth scene flow. During training, we aim to minimize the error between these estimates and the ground truth.

Let $\xi_k = [X_k,X_k+T_k]$ be the trajectory feature of an object
$\mathcal{O}_k$. $X_k$ and $X_k+T_k$ are the object centers at frame
$t$ and $t-1$, respectively.
Unless two objects have exactly the same object center and move with exactly the same translation trajectory, each $\xi_k$ is unique per object. Therefore, we can use it as a cue for motion-based, object segmentation. 

\section{Technical Approach}
\begin{figure*}[!htb]
\centering
\includegraphics[width=0.94\linewidth]{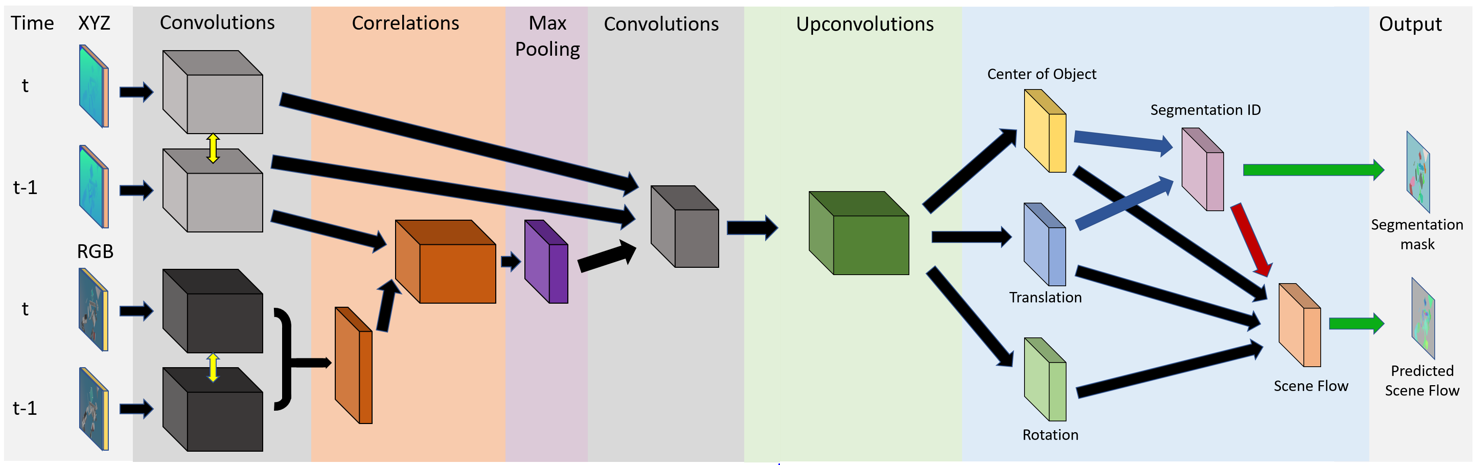}
\caption{Network architecture utilized in this paper. The RGB-D input
  is split into two components, RGB and XYZ, before being passed into
  Siamese neural networks. A correlation is performed on the output of
  the RGB Siamese network and applied to the XYZ features from time
  $t-1$. After a max pooling layer, the newly combined features undergo
  upconvolutions. The output of the upconvolutions is fed into 3
  different layers that predict the center of the object, translation,
  and rotation. Thereafter, the segmentation ID is determined using
  the center of the object and its predicted translation. For
  predicting scene flow, the translation, rotation, and input XYZ data
  is utilized. The final output is presented as a segmentation mask
  and scene flow predictions. Note that the blue, red, and green
  arrows do not have gradient flow.} 
\label{fig:sceneflownet}
\end{figure*}


\subsection{Rigid Motion and Object Scene Flow}
The first stage of the proposed model, displayed in
Fig.~\ref{fig:sceneflownet}, consists of two Siamese networks that
takes RGB images $\mathcal{I}^{t-1}$, $\mathcal{I}^{t}$ and point
clouds $\mathcal{P}^{t-1}$, $\mathcal{P}^{t}$ as inputs, each with
resolution $(W,H,3)$. The pair of point clouds is fed into the first
of these networks that outputs a new feature encoding denoted by
$\mathcal{P}f^{t-1}$ and $\mathcal{P}f^{t}$, respectively. We use the
VGG architecture~\cite{simonyan2014very} for this purpose. The shape
of the output feature is $(W/8,H/8,64).$ 

The pair of RGB images is fed into the second Siamese network that
outputs a new feature encoding denoted by $\mathcal{I}f^{t-1}$ and
$\mathcal{I}f^{t}$, respectively. We use the ResNet50 architecture and
its weights for initialization~\cite{he2016deep}. The shape of the
output feature tensor is $(W/8,H/8,256)$.  

The RGB image features are fed into a correlation layer similar to the
one used in FlowNetC~\cite{7410673}. A high correlation between
patches in consecutive RGB images indicates that they contain a
projection of the same physical object part. This correlation layer
parallels the brightness constancy assumption in traditional optical
and scene flow methods.  

\begin{figure}[htb]
\centering
\includegraphics[width=0.88\linewidth]{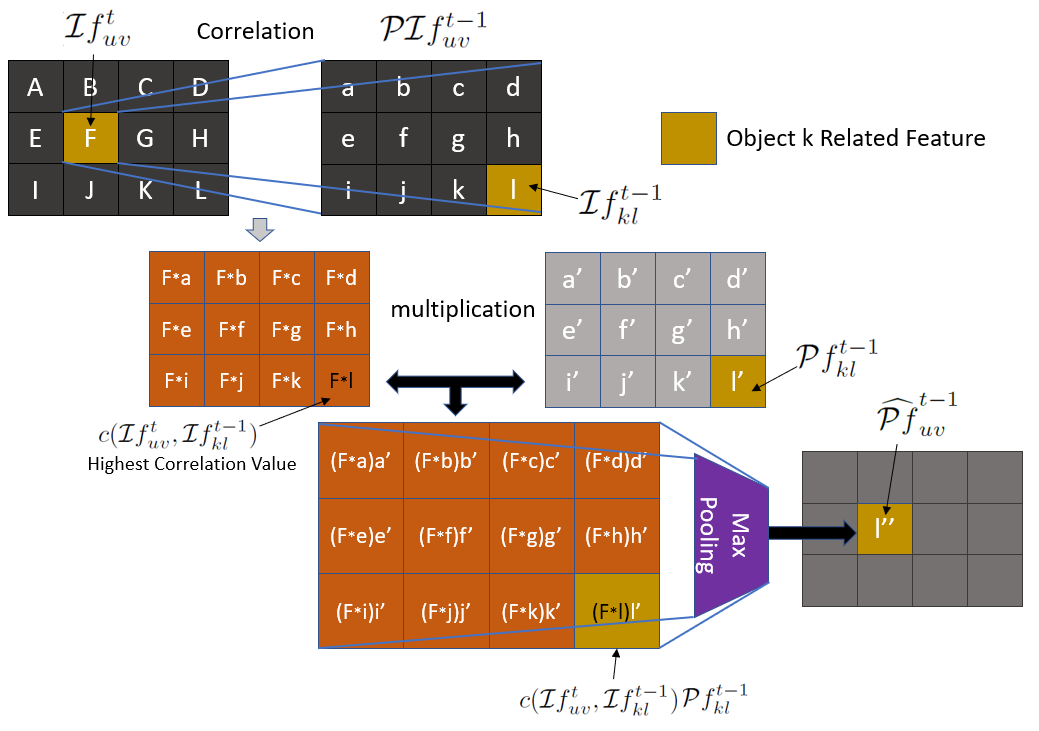}
\caption{Process of correlation and max-pooling. After two RGB feature
  maps (black) are generated, each cell in the feature map
  $\mathcal{I}f^{t}$ is correlated with every cell within a patch of
  the feature map $\mathcal{I}f^{t-1}$. Let us assume that the yellow
  cells $F$ and $l$ contain features corresponding to the same
  object. Therefore, their correlation $c(\mathcal{I}f^{t}_{F},
  \mathcal{I}f^{t-1}_{l})$ will be high.  These correlation values are
  used to weight corresponding cells of the XYZ feature map
  $\mathcal{P}f^{t-1}$ (gray). The result is fed into a max-pooling
  layer which in this example will output $c(\mathcal{I}f^{t}_{F},
  \mathcal{I}f^{t-1}_{l}) \mathcal{P}f^{t-1}_{l'}$. The final feature
  $l^{\prime\prime}$ containing object XYZ information at frame $t-1$
  will be placed at the same location as feature $F$ at frame $t$.} 
\label{fig:cc}
\end{figure}

Fig.~\ref{fig:cc} visualizes the correlation process encoded in the
layer. Let $\mathcal{I}f^{t}_{uv}$ denote a feature of RGB image
$\mathcal{I}^{t}$ at pixel $(u,v)$. Each feature is correlated with a
patch of features denoted by $\mathcal{P}\mathcal{I}f^{t-1}_{uv}$. The
patch is centered at $\mathcal{I}f^{t-1}_{uv}$ and has a side length
of $2L+1$, i.e. the dimension of the patch encoding is
$(2L+1,2L+1,256)$. The correlation operation between features
$\mathcal{I}f_{uv}^{t}$ and $\mathcal{I}f_{kl}^{t-1}$ inside the patch
$\mathcal{P}\mathcal{I}f^{t-1}_{uv}$ is defined as  
\begin{equation}
c(\mathcal{I}f^{t}_{uv}, \mathcal{I}f^{t-1}_{kl}) = \langle \mathcal{I}f^{t}_{uv},\mathcal{I}f^{t-1}_{kl} \rangle \text{ if } |u-k| \le L, |v -l| \le L
\end{equation}
The output vector of correlation between the single feature
$\mathcal{I}f_{uv}^{t}$ and corresponding patch
$\mathcal{P}\mathcal{I}f^{t-1}_{uv}$ has a dimension of
$(2L+1)^2$. The correlation is performed at each pixel within
$\mathcal{I}f^{t-1}$  with a stride of $(W/8,H/8)$. The final output
shape of the correlation layer is $(W/8,H/8,(2L+1)^2)$.  

Highly correlated RGB patches also indicate which parts in consecutive point clouds correspond to each other. 
We therefore multiply the correlation value tensor with the
corresponding $\mathcal{P}f^{t-1}$ features to get a weighted XYZ
feature encoding $\widehat{\mathcal{P}f}^{t-1}$. Then we apply max
pooling to this result along the feature dimension as follows: 
\begin{equation}
\widehat{{\mathcal{P}f}}^{t-1}_{uv} = \max_{\substack{|u-k| \le L \\ |v-l| \le L}}(c(\mathcal{I}f^{t}_{uv}, \mathcal{I}f^{t-1}_{kl})\mathcal{P}f^{t-1}_{kl} )
\end{equation}

We concatenate $[Pf^{t},Pf^{t-1},\widehat{Pf}^{t-1}]$ and feed this
into another encoder until reaching a feature map with size
(W/60,H/60,512) before feeding it into a decoder. Skip links are
created between encoder and decoder. The decoder generates three
images $\mathcal{Q}$, $\mathcal{T}$ and $\mathcal{X}$ representing
per-pixel estimates of rotation, translation and center position of
the object projected to that pixel. Per-pixel scene flow can then be
computed through Eq.~\ref{eq:sceneflow}. 



\subsection{Motion-based Segmentation}\label{sec:motion-based}
As defined previously, let $\xi_k = [X_k,X_k+T_k]$ represent the start
and end point of the object trajectory of $O_k$. Pixels
belonging to the same object $\mathcal{O}_k$ will have the same value
$\xi_k$. We assume that pixels belonging to different objects
have different values. Based on this we perform object segmentation. 

Our model makes a pixel-wise prediction $\hat{\xi}_{uv}$ of the
trajectory feature at pixel coordinates $(u,v)$. This is only an
approximation of the ground truth value. Therefore, each pixel
$(u,v)$ that corresponds to the same object $\mathcal O_k$ will predict
feature values $\hat{\xi}_{uv}$ that differ from ground
truth by some $\epsilon_{uv}$ such that $\hat{\xi}_{uv} = \xi_{uv} +
\epsilon_{uv}$.

To segment moving objects, we propose the
following inference process. Let $\mathcal{B}$ be an additional output
image of our model. A pixel at $(u,v)$ contains a scalar value
$B_{uv}$. This value is a radius estimate of the sphere that encloses
all pixels which belong to the 
same moving object, i.e. have a similar trajectory. The sphere is
centered at $\hat{\xi}_{uv}$. Any pixel at coordinates $(o,p)$ whose
$\hat{\xi}_{op}$ falls inside the sphere centered around
$\hat{\xi}_{uv}$ will be segmented as the same
object $\mathcal{O}_k$. Any pixel at $(m,n)$ whose $\hat{\xi}_{mn}$
falls outside the sphere will be part of the background or a different
object. In addition to $\mathcal{B}$, we also learn a mask
layer to discard pixel in this segmentation process that belong to the
background. 

To generate the ground truth of $\mathcal{B}^{gt}$, each pixel (u,v)
representing object $\mathcal{O}_k$ is annotated by half of the
minimum distance between $\xi_k$ and the trajectories
$\xi_l$ of all the other objects in the image pair: 
\begin{equation}
B^{gt}_{uv} = \frac{1}{2} \min_{k \ne l}\|\xi_k - \xi_l\|_2
\end{equation}

Inspired by region proposals~\citep{DBLP:journals/corr/RenHG015}, our
model also outputs an image denoted by $\eta$. Each pixel in this
image at $(u,v)$ contains the probability $\eta_{uv}$ that it is the
projection of the object centroid.  To generate the ground truth of $\eta$, we sort pixels representing object $\mathcal{O}_k$ by their distance to the object's centroid in ascending order. The top $D$ pixels per object in the input image
$\mathcal{I}$ will be labeled as 1, the rest will be labeled as
0. If the total number of pixels representing object $\mathcal{O}_k$
is less than $D$, all of them are labeled as 1. We found that $D=300$
worked well. This corresponds to approximately 10\% - 30\% of the
ground truth object pixels. The final performance is not very sensitive
to this parameter

\begin{figure}[t]
\centering
\includegraphics[width=0.9\linewidth]{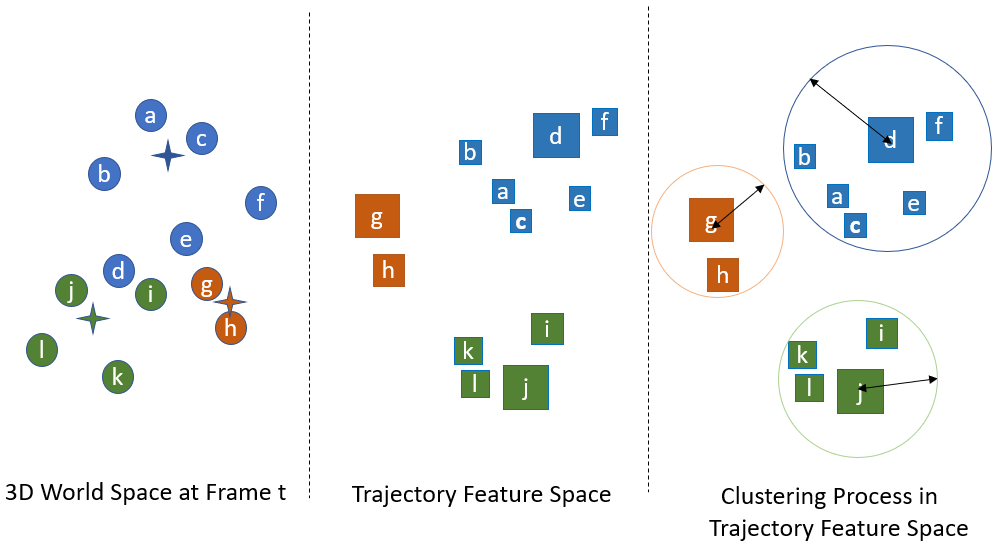}
\caption{Object segmentation process. Left: Points represent points in
  a point cloud. Stars represent ground-truth object centers. Same
  color indicates same object. Middle: Each square represents the
  trajectory features $\hat{\xi}$ in trajectory feature space each
  associated with a point on the left. The size of the
  squares represents the corresponding point's probability $\hat{\eta}$
  of being an
  object centroid. Right: The segmentation process cycles through the
  squares starting with those having the highest probability to be an
  object centroid. A sphere centered at one of those squares with
  radius $\hat{\mathcal{B}}$ then segments trajectories and corresponding points.} 
\label{fig:cluster}
\end{figure}

Given the predicted $\hat{\mathcal{B}}$ and $\hat{\eta}$, we can now
perform multi-object segmentation as visualized in
Fig.~\ref{fig:cluster}. Pixel $(u,v)$ with the maximum predicted probability
$\hat{\eta}_{uv}$ is proposed first. Given a sphere centered at
$\hat{\xi}_{uv}$ with radius $\hat{B}_{uv}$, all pixels $(m,n)$ with a
trajectory $\hat{\xi}_{mn}$ enclosed by this sphere are assigned to
object $O_1$. All pixel assigned to $O_1$ are removed from the set of
unsegmented pixels before segmenting the next object. The remaining
pixel at $(o,p)$ with the highest $\hat{\eta}_{op}$ is used as the seed for
segmenting $O_2$.  This process is repeated until all foreground
pixels are assigned an object id $k$. The final object translation
$T_k$ and rotation $R_k$ 
is computed by averaging over all pixels with the same id. Based on
this, also the scene flow can be recomputed. 

\subsection{Loss Function} \label{sec:loss}
We use the following training loss:
\begin{align}
L = & \lambda_{m}L_{m}  +\lambda_{center}L_{center} + L_{p} \nonumber\\
 & +\lambda_{var}L_{var}+\lambda_{vio}L_{vio}.
\end{align}
In the following, we define each term. Note that all pixel-wise loss
terms $L_p$, $L_{center}$, $L_{var}$ and $L_{vio}$ are only computed
on the ground truth foreground pixel.  

\subsubsection{Mask Loss}
$L_{m}$ is the cross entropy loss between the ground truth and
estimated foreground/background segmentation. If a pixel is the
projection of an object point, we assign 1 as ground truth; otherwise
0. 

\subsubsection{Cluster Center Loss}
Cross-entropy loss $L_{center}$ is used to learn the probability
$\eta_{uv}$ of a pixel $(u,v)$ to be the object center as described in
Sec.~\ref{sec:motion-based}. 

\subsubsection{Pixel-wise Loss}
We use a pixel-wise loss $L_{p}$ on the predicted object rotation
$Q_{uv}$, translation $T_{uv}$, scene flow
$S_{uv}$, enclosing sphere radius $B_{uv}$ and trajectory $\xi =
[X_{uv},X_{uv}+T_{uv}]$. For each attribute, we use the L2-norm to measure
and minimize the error between predictions and ground truth. Note that
the loss on each attribute is also differently weighted. We denote
their corresponding weights $\lambda_{Q}$, $\lambda_{T}$, $\lambda_X$,
$\lambda_{S}$, $\lambda_B$ and $\lambda_{\xi}$. 

\subsubsection{Variance Loss}
We use $L_{var}$ to encourage pixels $(u,v)$ belonging to the same
object $\mathcal{O}_{k}$ to have similar trajectories $\xi_{uv}$ and
thereby to reduce their variance.  
\begin{equation}
L_{var} = \sum_{k}\frac{1}{N_{k}}\sum_{(u,v) \in \mathcal{O}_k}\|
\hat{\xi}_{uv} - \bar{\hat{\xi}}_{uv} \|^2 
\end{equation}

where $\bar{\hat{\xi}}_{uv}$ is the mean value of $\hat{\xi}_{uv}$
over all $N_k$ pixels belonging to $\mathcal{O}_k$.  

\subsubsection{Violation Loss}
$L_{vio}$ penalizes pixels $(u,v)$ that are not correctly
segmented. Any predicted trajectory $\hat{\xi}_{uv}$ that is more than
$\frac{1}{5}\mathcal{B}_{uv}$ away from the ground truth $\xi_{uv}$
will be pushed towards the ground truth trajectory by the violation
loss. Note that $\mathcal{B}_{uv}$ refers to the radius of enclosing sphere. 

\begin{equation}
L_{vio} =\sum_{k}\sum_{(u,v)\in \mathcal{O}_k}
\mathbf{1}\{\|\hat{\xi}_{uv} -\xi_{uv}\|_2 >
\frac{1}{5}\mathcal{B}_{uv}\} \|\hat{\xi}_{uv} -\xi_{uv}\|_2 \nonumber
\\ 
\end{equation}

The variance and violation loss are designed to train the clustering framework described in
~\ref{sec:motion-based}.  

\section{Dataset}\label{sec:dataset}
We generated a new dataset that consists of RGB-D image pairs showing
dynamic scenes. These scenes contain a large variety of rigidly-moving objects.
See Fig.~\ref{fig::synthetic_table} for some example frames. 
To ensure a diverse data set, we used 31594 3D object mesh models from
ShapeNet~\citep{shapenet2015} covering 28 categories. We split these
models into a training, validation and test set with 21899, 3186  and
6509 objects respectively. Model sizes are adjusted to simulate their real
world sizes~\cite{lin2017crossmodal}. For each scene, 1-30 object
models are randomly selected. For simulating realistic object motion,
we use Bullet~\citep{Coumans:2015:BPS:2776880.2792704} as physics
engine. The objects are put close to each other at 0.2 meter above a
surface. After simulation begins, they start to fall down to the surface 
and collide with each other in the process. The RGB-D camera is static and the simulation runs at 60 Hz. We extract frame 20 and 80 from the simulated image sequence as RGB-D image pair. They are 1 second apart with 
an average object displacement of 0.085 meters. We synthesize 24994,
3360 and 7186 frame pairs for training, validation and test set. Note
that the object models are not re-used across these data sets. In
total, we generated 35540 pairs of 
consecutive RGB-D frames using Blensor~\cite{gschwandtner2011blensor}
to ensure realistic depth data. For each rendered RGB image pair, we
randomly sample an image from the SUN397 dataset~\cite{xiao2010sun} to
simulate textured floor or we use a single color. We also randomly
change the lighting conditions (number of light sources, their
positions and energies) and camera viewpoint. We do not add artificial
noise in the raw dataset for two reasons. Different sensors like
time-of-flight or structured light have different noise
patterns. Adding one type of noise pattern into the dataset might
increase the simulation-to-reality gap when other sensors are
used. Extra noise can be dynamically added into the neural network
training procedure as data augmentation procedure.  

\subsection*{Annotating Objects with Rotational Symmetry}
\begin{figure}[t]
\centering
\includegraphics[width=0.8\linewidth]{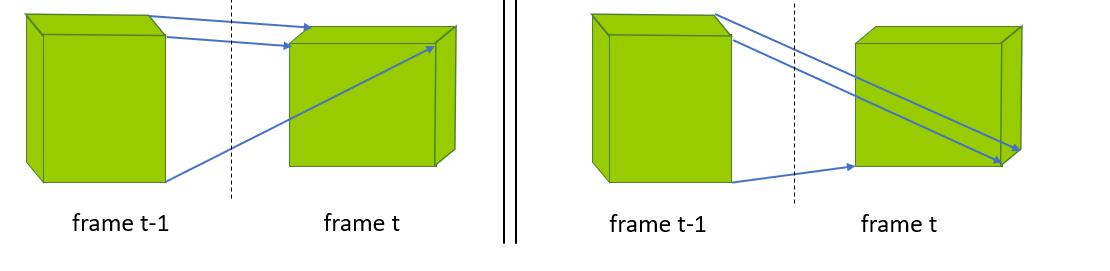}
\caption{A rotationally symmetric object for which two different rotations yield the same RGB-D data. Therefore, multiple solutions for scene flow exist.}
\label{fig:rotational symmetry}
\end{figure}

Some of the objects in ShapeNet~\citep{shapenet2015} are rotationally symmetric, e.g. bottles and bowls. Rotational symmetry is a common object attribute especially for human-made objects. However, the rotation of such an object around its symmetry axis is not observable in an image pair (especially when uniformly colored) as there might be multiple or even infinite solutions.
There are different orders of rotational symmetry denoted by $C_2$, $\cdots$, $C_n$, $\cdots$,$C_{\infty}$.  An object with $C_n$ means that it will remain the same after rotating about the rotation axis by $\pm 360/n$ degrees. An object might contain several different rotational symmetries. Fig.~\ref{fig:rotational symmetry} illustrates an example.

This has implications for the ground truth annotation of our dataset. If we directly use the ground truth rotation provided by the simulator, the network might not converge during training as more than one rotation might lead to the same RGB-D data.
In the following, we describe a procedure to map the ground truth rotation of an object about its symmetry axis to the rotation with minimum angular displacement.
Consider an object with $C_n$ rotational symmetry. Let $\bar r_{t-1}$ and $\bar r_{t}$ denote this axis of symmetry at frame $t-1$ and frame $t$, respectively. Let the rotation provided by the simulator be given as a quaternion $q = [q_0, q_x, q_y, q_z]^T$. We decompose the rotation $q$ into a rotation $\alpha$ about $C_n (\bar{r}_{t-1})$ and a rotation $\theta$ perpendicular to $C_n: ~(\bar{r}_{\bot} = \bar{r}_{t-1} \times \bar{r}_t)$.
\begin{align}
\alpha = & 2\tan^{-1}(\frac{r_{t-1,x}q_x+r_{t-1,z}q_z+r_{t-1,z}q_z}{q_0})
\label{eq::quat_decomp1}\\
\theta = & 2\cos^{-1}(q_0/cos(\frac{\alpha}{2}))
\label{eq::quat_decomp2}
\end{align} 
$\alpha$ is then adjusted to be $\hat\alpha \in (-\pi/n,\pi/n]$. 
This corresponds to the minimum angular displacement leading to the same observation as the original angle. From this, we can construct a new quaternion $\hat q$ which corresponds to the rotation of $\hat \alpha$ about $\bar{r}_{t-1}$ and rotation of $\theta$ about $\bar{r}_{\bot}$. Note that if $\alpha = \hat\alpha \in (-\pi/n,\pi/n]$ then $q = \hat q$. This operation is performed on all the rotational axis of symmetry. 
With this procedure, we reduced the ambiguous cases to a very small number, e.g. to uniformly-colored objects with non-orthogonal axes of symmetry (of which there exists one among our models) or rotations as shown in Fig.~\ref{fig:rotational symmetry} where the minimum angular displacement can either refer to a rotation in the positive or negative direction.


\section{Experiments}
We report the performance of the proposed model quantitatively on the synthesized dataset and qualitatively on real
data. We evaluate accuracy in scene flow prediction by comparing to
PD-Flow~\citep{jaimez2015primal}, semi-rigid scene flow (SR-Flow)~\cite{quiroga2014dense} and \citet{7989459}. We evaluate motion-based
segmentation performance by comparing to Higher-Order Minimum Cost
Lifted Multicuts (HOMC)~\citep{keuper2017higher}. 

Furthermore, we compare to variants of the proposed architecture. We refer to the network in Fig.~\ref{fig:sceneflownet} as $\textbf{OurC}$ and propose a simpler neural net architecture denoted by $\textbf{OurS}$. It concatenates all four input images and feeds it into the encoder. Most importantly, it drops the correlation and max pooling layer. The remaining model architecture is the same. 
$\textbf{OurC+vL}$ denotes added variance and violation loss compared
to training $\textbf{OurC}$. Our model $\textbf{OurC+vL}$
simultaneously predicts pixel-level segmentation IDs and scene
flow. Given all pixels with the same, predicted object ID, we compute
the mean object center $\bar{X}_k$, translation $\bar{T}_k$ and
rotation $\bar{Q}_k$. $\textbf{OurC+vL+Rig}$ denotes the model with
added rigidity constraints for improved scene flow estimation. After we infer the segmentation mask per object, we average the rigid transformations from pixels predicted to represent the same object. The rigid transformation is then applied to pixels to recalculate scene flow.

We conduct our experiments on an NVIDIA P100 with TensorFlow. For
training, we use the Adam optimizer~\citep{kingma2014adam} with its
suggested default parameters of $\beta1=0.9$ and
$\beta2=0.999$ along with a learning rate $\alpha=0.0001$~\cite{kingma2014adam}. We use a batch size of 12 image pairs. The input RGB-D images have a resolution of $240\times 320$. The loss weights, as defined in Sec.~\ref{sec:loss}, are set to $\lambda_m=1.0$, $\lambda_{var}=0.1$,
$\lambda_{vio}=0.1$, $\lambda_{Q}$=0.1, $\lambda_{T}$=100.0,
$\lambda_X$=10.0, $\lambda_{S}$=10.0, $\lambda_B$=1.0 and
$\lambda_{\xi}$=1.0. 

\subsection{Evaluation of Scene Flow Performance}
We compare the proposed method with the aforementioned approaches 
using standard evaluation metrics as defined in~\citep{survey}: end point error (EPE) and $4D$ average angular error (AAE) error. Each metric is calculated as averages over the entire
image and is reported in cm and degrees, respectively. Because it
is impossible to calculate the scene flow for an object that is only
present in one of the two frames, we also report masked EPE and masked
AAE which calculates the desired metrics only on objects that are in
both frames. The results are presented in
Fig.~\ref{fig::res_scene_flow}.


\begin{figure}[t]
\tiny
\renewcommand\arraystretch{1.7}
\centering
\begin{tabular}{p{0.15\linewidth} ||P{0.11\linewidth} P{0.11\linewidth} ||P{0.11\linewidth} P{0.11\linewidth} 
|| P{0.08\linewidth}}
\hline
\hline
\scriptsize Method & \multicolumn{2}{c||}{\scriptsize  EPE in cm} & \multicolumn{2}{c||}{\scriptsize  AAE in degrees}  &\scriptsize  Runtime \\
 &\scriptsize all & \scriptsize masked &\scriptsize all &\scriptsize masked &\scriptsize seconds\\
 \hline
 PD-Flow\cite{jaimez2015primal}& 2.830 $\pm$4.23& 8.041 $\pm$5.10& 1.607 $\pm$2.38& 4.572 $\pm$2.87&\textbf{0.046} \\
SR-Flow \cite{quiroga2014dense}& 2.040 $\pm$3.77 & 6.859 $\pm$4.69 &1.155 $\pm$2.12 & 3.898 $\pm$2.63& $\ge$1\\ 
Jaimez et al. \cite{7989459}& 2.330 $\pm$3.94 & 6.431 $\pm$5.20& 1.317 $\pm$2.20& 3.643 $\pm$2.91& 0.083 \\
 \hline
 OurS & 1.643 $\pm$3.07 & 5.324 $\pm$3.56 & 0.928 $\pm$1.72 & 3.020 $\pm$2.00& 0.059\\
 OurC & 1.330 $\pm$2.60 & 4.333 $\pm$2.92 & 0.750 $\pm$1.45 &2.457 $\pm$1.64& 0.078\\
 OurC+vL & 1.315 $\pm$2.57 & 4.333 $\pm$2.95 & 0.742 $\pm$1.44 & 2.457 $\pm$1.66& 0.078\\
 OurC + vL + Rig & \textbf{1.303} $\pm$2.55 & \textbf{4.290} $\pm$2.93 & \textbf{0.734} $\pm$1.43 & \textbf{2.432} $\pm$1.64& 0.121\\
\hline
\hline
\end{tabular}
\caption{Performance of scene flow prediction measured in endpoint error (EPE) and average angular error (AAE) with standard deviation. {\em Masked\/} only contains data from objects that appear in both frames. The learned models outperform the baselines in terms of mean error and std  \textbf{OurC} and its variants perform better than the simple model version \textbf{OurS} without the correlation layer.}
\label{fig::res_scene_flow}
\end{figure}
All our proposed models outperform the aforementioned approaches both in mean and standard deviation. Furthermore, \textbf{OurC} and its variants perform better than the simple
model version \textbf{OurS} without the correlation layer. This comes
at the expense of a higher processing time. However, the most complex
model $\textbf{OurC+vL+Rig}$ can still run at 8.3 frames per
second. 

\subsection{Evaluation of Motion-based Segmentation} 
We evaluate our model's ability to perform motion-based segmentation
by comparing to HOMC, the state-of-the-art technique
by \citet{keuper2017higher}. This method requires a sequence of RGB
images. To satisfy the input requirement of HOMC, we generate an additional test dataset called \textbf{TestSeq} following the procedure outlined in Sec.~\ref{sec:dataset}. In \textbf{TestSeq} there are 1302 image sequences each consisting of 8 frames with indices 20, 30, 40, 50, 60, 70, 80 and 90. The original data set only has 2 frames, frame 20 and 80. From this we create the data set \textbf{Test} in which these two frames are repeated 5 times (20, 80, 20, 80, 20, 80, 20, 80, 20, 80) such that it can serve as input to HOMC. \citet{keuper2017higher} provided an executable file upon
request. We run HOMC\cite{keuper2017higher} on the sequences with a subsampling of 4 and a prior cut probability of 0.5. For our proposed models, frames 20 and 80 compose the input image pair for all experiments. 


To evaluate the segmentation results produced by HOMC and our three network
variants, we rely on four metrics that are frequently used in
segmentation papers: precision, recall, F-measure, and extracted
objects~\citep{KB15b}. We compute the metrics on the segmentation of frame 80 by following the
convention in~\citep{ochs2014segmentation}. We use an F-measure threshold of
0.75. The results are reported
in Fig.~\ref{tab::seq_motion_seg}.

\begin{figure}[t]
\tiny
\renewcommand\arraystretch{1.7}
\centering
\begin{tabular}{p{0.08\linewidth} ||P{0.03\linewidth} P{0.055\linewidth} ||P{0.03\linewidth} P{0.055\linewidth} 
||P{0.03\linewidth} P{0.055\linewidth}
||P{0.07\linewidth} P{0.08\linewidth}
}
\hline
\hline
\scriptsize Method & \multicolumn{2}{c||}{\scriptsize Precision} & \multicolumn{2}{c||}{\scriptsize Recall} &  \multicolumn{2}{c||}{\scriptsize F-measure} & \multicolumn{2}{c}{\scriptsize Extracted Objects}\\
&\fontsize{5.5}{7.2}\selectfont Test &\fontsize{5.5}{7.2}\selectfont TestSeq &\fontsize{5.5}{7.2}\selectfont Test &\fontsize{5.5}{7.2}\selectfont TestSeq &\fontsize{5.5}{7.2}\selectfont Test &\fontsize{5.5}{7.2}\selectfont TestSeq &\fontsize{5.5}{7.2}\selectfont Test &\fontsize{5.5}{7.2}\selectfont TestSeq\\
\hline
HOMC\cite{keuper2017higher}& \textbf{0.833} &\textbf{0.797} & 0.195 & 0.282 & 0.111& 0.186  & 3909/64556 & 1172/11782 \\
\hline 
OurS & 0.697 & 0.714 & 0.735 & 0.749 & 0.671 & 0.683& 36369/64556 & 7077/11782 \\
OurC & 0.756 & 0.763 & 0.757 & 0.771 &0.696 & 0.711& 41274/64556 & 7839/11782 \\
OurC+vL & 0.766 & 0.768 & \textbf{0.787} & \textbf{0.783} & \textbf{0.730}& \textbf{0.725} & \textbf{43719/64556} & \textbf{8015/11782}\\
\hline
\end{tabular}

\caption{Performance of motion-based segmentation. \textbf{Test} refers to the dataset containing repetitions of 2 image frames. \textbf{TestSeq} refers to the dataset containing 8 image frames. The proposed models significantly outperform HOMC that requires longer image sequences but cannot rely on the strong depth cue. The performance increase of $\textbf{OurC+vL}$ over the simpler models highlights the importance of the correlation layer and the variance and violation loss.}
\label{tab::seq_motion_seg}
\end{figure}

\begin{figure*}[htb!]
\centering\includegraphics[width=0.86\linewidth]{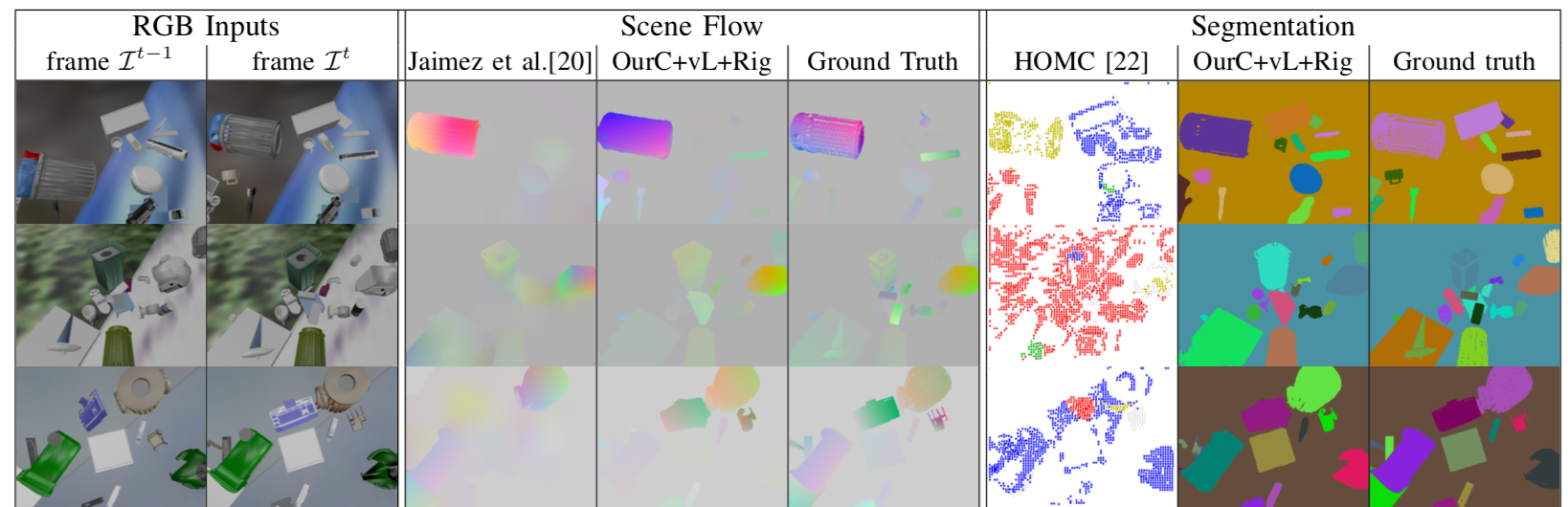}
\caption{Performance comparison of the proposed method on our synthetic data set. First two columns show RGB inputs, next columns are Scene Flow from Jaimez et al.~\cite{7989459}, our method, and the ground truth, the final three columns correspond to segmentation results from HOMC \cite{keuper2017higher}, our method, and the ground truth. In scene flow images, green, blue, red intensities are proportional to the velocities along $X,~Y,~Z$ respectively. In the HOMC segmentation, colored pixels (not gray or white) have been successfully clustered with longer trajectories to produce valid segmentations.}
\label{fig::synthetic_table}
\end{figure*}

On both datasets, HOMC~\citep{keuper2017higher} achieves high precision and low recall values indicating undersegmentation. In \textbf{TestSeq}, HOMC extracts more objects with
higher recall and F-measure scores than \textbf{Test}, emphasizing the dependence on an actual image sequence.  All our proposed methods show a significant improvement on the recall, F-measure, and extracted
objects metrics while retaining a high precision score. While
HOMC relies on a longer sequence of images and takes more than 30 seconds to process the sequence, it does not require depth information.
A few example results are displayed in
Fig.~\ref{fig::synthetic_table}. These results highlight another
advantage of our approach, that the resulting segmentation is dense.

\subsection{Architecture Design Analysis}
\subsubsection{Effects of correlation layer}
\begin{figure}[t]
\centering\includegraphics[width=0.5\linewidth]{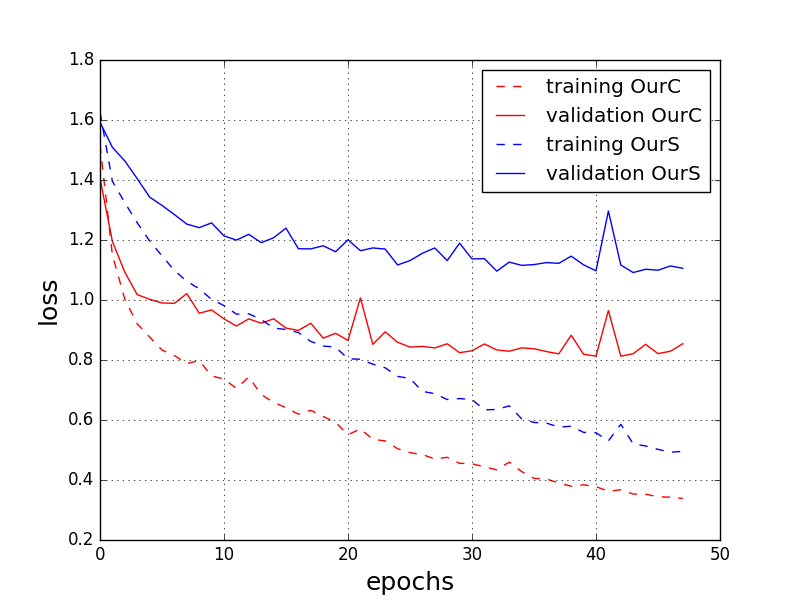}
\caption{Loss curve at each epoch during training and validation for
our model with and without the correlation layer: $\textbf{OurC}$ and
$\textbf{OurS}$} 
\label{fig::loss}
\end{figure}

We report the training and validation loss curve in
Fig.~\ref{fig::loss}. $\textbf{OurC}$ has a much lower
training and validation loss than $\textbf{OurS}$. We also showed that
$\textbf{OurC}$ outperforms $\textbf{OurS}$ both in scene flow
prediction and motion-based segmentation. This demonstrates the impact of adding a correlation layer in $\textbf{OurC}$. It forces our model to learn the similarity between consecutive RGB features which makes $\textbf{OurC}$ more robust to changes such as lighting conditions or viewpoints. 

\subsubsection{Effects of using variance and violation loss}
We utilize the variance loss to reduce the statistical variance of predicted trajectory features. The violation loss penalizes outliers in the training process. Compared to $\textbf{OurC}$, $\textbf{OurC+vL}$ improves motion-based segmentation, but only leads to small improvements on scene flow prediction. 

\subsubsection{Effects of using rigid motion cues}
The best scene flow prediction performance is achieved by adding rigid
constraints ($\textbf{OurC+vL+Rig}$). However the improvement over
$\textbf{OurC}$ is only marginal. The difference to $\textbf{OurS}$
remains significant, underlining the importance of correlation
layer. 
  
\subsection{Results and Analysis on Real World Data}
Finally, we demonstrate the networks ability to perform in a real
world setting. We recorded real RGB-D data with the Intel RealSense
SR300 Camera. The data includes large displacements, occlusions, and
collisions. It was captured using a diverse set of objects with
varying geometries, textures, and colors.  Note that we do not have
any ground truth annotations and that the model is not fine-tuned to
transfer from synthetic to real data.   

\begin{figure*}[htb!]
\centering\includegraphics[width=0.86\linewidth]{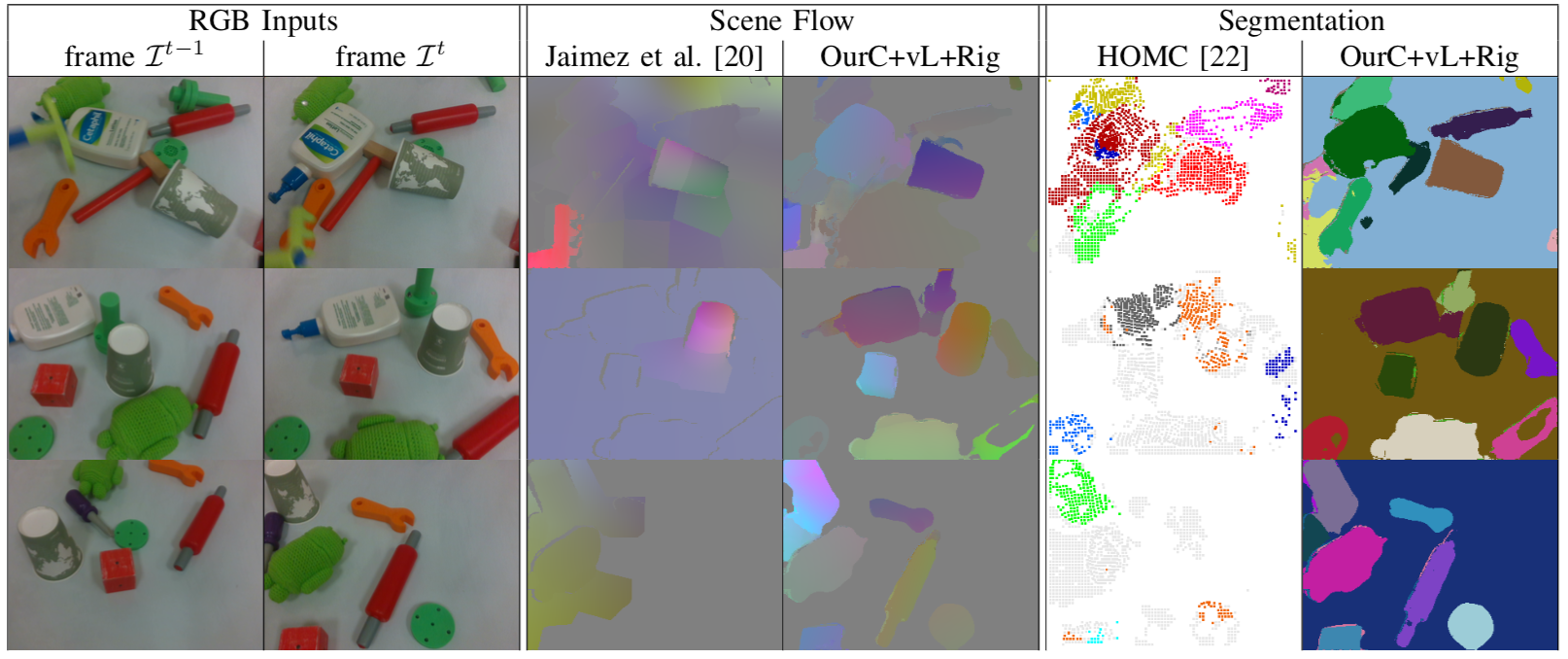}
\caption{Performance comparison of the proposed method on the real-world data set. First two columns show RGB inputs, next columns are Scene Flow from Jaimez et al.~\cite{7989459} and our methods, last two columns correspond to segmentation results from HOMC \cite{keuper2017higher} and our method.}
\label{fig::realworld_table}
\end{figure*}

We apply HOMC~\citep{keuper2017higher} on the stream of real data as one long sequence. We use our $\textbf{OurC+vL+Rig}$
model to process real data sequences. Every pair of consecutive images forms one image pair which are fed into our neural network. 
Some example images and corresponding outputs are displayed in
Fig.~\ref{fig::realworld_table}. The accurate real world segmentation
and scene flow prediction results strongly indicate the small
sim-to-real transfer gap of the proposed model. 

There are still some failure cases including inaccurate object boundaries due to noisy sensor data and false positive segmentations due to varied lighting conditions. Also if two objects are moving along extremely similar trajectories, it is difficult to segment them. 
This could be potentially alleviated by concatentating rotational motion to the trajectory feature.
Other limitations of our method include: inability to generalize to non-flat surfaces or non-rigid objects. The generalization problem of learning-based methods could be mitigated by transfer-learning techniques e.g.~\cite{ganin2015unsupervised}.
\renewcommand{\tabcolsep}{0pt}

\renewcommand{\tabcolsep}{0pt}
\newcommand*\rot{\rotatebox{90}}

\section{Conclusion} 
We proposed a deep neural network architecture that given two consecutive RGB-D images can accurately estimate object scene flow and motion-based object segmentation. We demonstrated this on a new and challenging, synthetic data set that contains a large variety of graspable objects moving simultaneously. We showed that the correlation layer makes a crucial difference to training time and accuracy and outperforms state of the art baselines in scene flow prediction and motion-based segmentation. Additionally, we showed how our approach performs on real RGB-D data when only trained on synthetic data. The results look qualitatively more accurate than baseline methods. Overall, we demonstrated the power of learning based methods over traditional methods in situations of large displacements and strong occlusions. 
In future work, we will explore how this approach enables agile, robotic manipulation in cluttered scenes. 



\footnotesize{
\bibliographystyle{plainnat}
\bibliography{references}
}
\end{document}